\documentclass[12pt]{article}
\usepackage{fontspec}

\newfontfamily\ipafont{CharisSIL-Regular.ttf}[
  BoldFont={CharisSIL-Bold.ttf},
  ItalicFont={CharisSIL-Italic.ttf},
  BoldItalicFont={CharisSIL-BoldItalic.ttf},
]

\newfontfamily\cjkfont{HaranoAjiMincho-Regular.otf}[
  BoldFont={HaranoAjiMincho-Bold.otf},
]

\usepackage[
  letterpaper,
  inner=1.25in,
  outer=1in,
  top=1in,
  bottom=1.25in,
  marginparwidth=0.6in,
]{geometry}

\usepackage[british]{babel} 
\usepackage[final]{microtype}

\usepackage{titlesec}

\titleformat{\section}
  {\normalfont\scshape}
  {\llap{\thesection\quad}}
  {0pt}
  {}

\titleformat{\subsection}
  {\normalfont\normalsize\scshape}
  {\thesubsection\quad}
  {0pt}
  {}

\titleformat{\subsubsection}
  {\normalfont}
  {\thesubsubsection\quad}
  {0pt}
  {}

\titlespacing*{\section}{0pt}{2ex plus 1ex minus .2ex}{1ex plus .2ex}
\titlespacing*{\subsection}{0pt}{1.5ex plus 1ex minus .2ex}{0.5ex plus .2ex}
\titlespacing*{\subsubsection}{0pt}{1ex plus 0.5ex minus .1ex}{0.3ex plus .1ex}

\usepackage{fancyhdr}
\renewcommand{\sectionmark}[1]{%
  \markboth{\ifnum\value{secnumdepth}>0 \thesection\quad\fi #1}{}%
}
\usepackage{xcolor}
\definecolor{linkmaroon}{RGB}{128, 0, 32}

\usepackage{hyperref}
\hypersetup{
  colorlinks=true,
  linkcolor=linkmaroon,
  citecolor=linkmaroon,
  urlcolor=linkmaroon,
  pdfauthor={Brett Reynolds},
}

\usepackage{orcidlink}

\usepackage[style=american]{csquotes} 
\MakeOuterQuote{"}

\newcommand{\term}[1]{\textsc{#1}}

\newcommand{\mention}[1]{\textit{#1}}

\usepackage{marvosym}

\usepackage{amsmath,amssymb}

\usepackage{langsci-gb4e}
\makeatletter
\@ifundefined{noautomath}{}{\noautomath}
\makeatother

\usepackage{enumitem}
\setlist{noitemsep, leftmargin=*}
\setlist[enumerate]{label=\arabic*.}
\setlist[itemize]{label=--}

\usepackage{booktabs}
\usepackage{array}
\usepackage{graphicx}
\graphicspath{{figures/}}

\usepackage[backend=biber,style=apa,natbib=true,doi=true,isbn=false,url=true]{biblatex}
\IfFileExists{references-standalone.bib}
  {\addbibresource{references-standalone.bib}}
  {\addbibresource{references.bib}\IfFileExists{references-local.bib}{\addbibresource{references-local.bib}}{}}

\usepackage{xspace}

\newcommand{\aidisclosure}[1]{%
  \par\medskip
  \noindent{\small\textsc{AI use.}\enspace The large language models #1 served as
  drafting and editing aids throughout the preparation of this paper. I am
  responsible for all theoretical claims, arguments, errors, and interpretive
  choices.\par}%
  \medskip
}

\usepackage{float}
\usepackage{tikz}
\usepackage{caption}
\usetikzlibrary{arrows.meta, positioning}
\hypersetup{
  pdftitle={When Benchmark Inferences Do Not Compose: Projectibility in AI Evaluation},
  pdfkeywords={AI evaluation, benchmark validity, projectibility, construct validity, generalization, AGI}
}
\newcommand{\instab}{\mathrm{INS}}
\newcommand{\wtd}{\mathrm{WTD}}
\newcommand{\Warr}{\operatorname{Warr}}

\newif\ifanon
\anonfalse
\ifanon\hypersetup{pdfauthor={}}\fi

\title{When Benchmark Inferences Do Not Compose:\\Projectibility in AI Evaluation}
\ifanon
  \author{}
\else
  \author{Brett Reynolds \orcidlink{0000-0003-0073-7195}\thanks{\href{mailto:brett.reynolds@humber.ca}{brett.reynolds@humber.ca}}\\
  Humber Polytechnic \& University of Toronto}
\fi
\date{}

\begin{document}

\maketitle

\begin{abstract}
An AI benchmark result rarely reaches a consequential claim in one step. Evaluators generalize it to further cases, interpret it as evidence of capability, extrapolate it to new tasks, transport it to another system or site, and combine it with assumptions about human review and downstream consequences. Validity-centred approaches require evidence for each claim. This paper makes explicit and operationalizes a problem those approaches leave to the analyst: warranted links don't automatically make a warranted chain. The target of one study may not be the source of the next; system, population, outcome, or conditions may change at the interface; and shared data or model lineage may make apparently independent support dependent.

\term{Projectibility} concerns whether a bounded extension from observed to unobserved cases is warranted. Goodman supplies the problem of rival extensions; argument-based validity supplies an architecture for testing them. The contribution is an interface audit for distributed AI evidence: typed source and target descriptions, and a procedure separating endpoints that never meet from endpoints that meet while warrant fails to cross. A legal-research case shows how benchmark evidence and a deployment study can each be sound while remaining parallel. A known-truth demonstration shows why aggregate stability can erase distinctions a later projection requires. The resulting \term{projectibility audit} is a link-by-link record of what each step projects and on what evidence, and it diagnoses unsupported joins in benchmark-to-use arguments.
\end{abstract}


\aidisclosure{ChatGPT 5; Claude Sonnet 4.5 and Opus 4.8; Gemini 3.1 Pro}

\section{Composition as a Distinct Problem}\label{sec:introduction}

A benchmark result rarely supports a consequential claim by itself. Consider a common chain of reasoning. A base model scores highly on a multidomain battery. The score is interpreted as evidence of general reasoning. A product built from that model is expected to perform a professional task. Human review is expected to improve the final work. That expectation, projected savings, and an error tolerance support authorization.

The chain spans four empirical objects: benchmark responses, outputs from a tool-using application, work after human review, and consequences under a policy. A capability attribution concerns the base model and usually licenses the move from benchmark to application performance. The score records only the benchmark responses, not the other three objects.

Psychological and educational measurement provide a disciplined response to this problem. On an argument-based view, validity belongs to a proposed interpretation and use of scores, not to a test considered apart from any claim \citep{messick1995Validity,kane2013validating}.

Recent work brings that view into AI evaluation, mapping benchmark results to the particular claims and decisions they're meant to support \citep{salaudeen2025measurement,bean2025measuring,freieslebenZezulka2025benchmarking}. These approaches substantially improve on treating a benchmark as valid or invalid without specifying what it's supposed to warrant.

They also make a further problem visible. Evaluation arguments are usually chains, but support for their links doesn't automatically compose. A benchmark may support a claim about new items generated under its rules. A firm may separately show that a particular human--AI procedure works on a sample of its own cases. Both results can be warranted without the first supplying a premise for the second. The local study may bypass the benchmark rather than extend it.

Conversely, two studies may appear to meet at a shared noun such as \mention{legal reasoning}, \mention{draft quality}, or \mention{human oversight} while operationalizing different objects, populations, and outcomes. In either situation, adding the studies together doesn't produce the broad claim. These aren't only possibilities. Legal-research vendors advertised retrieval-grounded tools as hallucination-free on the strength of how those tools were built, and a preregistered evaluation later measured hallucination on more than 17\% of queries \citep{magesh2025legalhallucination}.

The problem isn't that machine outputs are fallible while human judgment isn't. Human judgments and institutions also fail. AI makes the interface problem especially pressing because one evaluated component can be replicated at high volume, modified into many applications, and inserted into workflows whose relevant evidence is distributed among different actors. An unsupported bridge can become a repeated, correlated failure while no single study observes the complete chain.

The failure mode has a compact statement, a non-composition principle:
\begin{equation}\label{eq:noncomposition}
\Warr(C_1)\ \land\ \Warr(C_2)
\ \not\Rightarrow\
\Warr(C_2\!\circ C_1).
\end{equation}
In words: warrant for \(C_1\) and warrant for \(C_2\), each established separately, don't yield warrant for the composed inference \(C_2\!\circ C_1\), which runs \(C_1\) first and then \(C_2\). Here \(C_1\) and \(C_2\) are links presented as adjacent. If \(C_1\)'s target isn't \(C_2\)'s source, the composition isn't well-typed as stated and a bridge is owed; Equation~\ref{eq:noncomposition} covers defined composition whose warrant fails to transmit. Equation~\ref{eq:noncomposition} states a non-composition principle, not a theory of warrant.

Writing \(\Warr\) as a predicate is a notational convenience: \(\Warr(C)\) means warrant sufficient for the declared use. This threshold notation doesn't imply that support for a fixed claim is binary or reducible to one scale. Warranted composition requires alignment, or a separately warranted bridge, in object, population, conditions, outcome, and period; compatible assumptions and effect modifiers; and propagation of dependence and uncertainty.

The term \term{projectibility} comes from Goodman's treatment of induction \citeyearpar{goodman1983}, but the account developed here doesn't offer a general solution to induction and doesn't make entrenchment a sufficient condition. It gives a place within a validity argument for a narrower question: exactly what is being projected, across which boundary, under which assumptions, and with what evidence against the differences that could defeat the extension? A \term{projectibility audit} records the answer link by link and diagnoses unsupported joins rather than treating separately warranted links as automatically composable.

This focus separates the paper from neighbouring proposals without displacing them. \term{Nomological networks}, the relations a construct is predicted to bear to tasks, processes, and external criteria, along with generalizability designs, estimand specifications, and selection diagrams, each supply evidence or structure for a projection \citep{cronbach1955construct,freiesleben2026establishing,brennan2001generalizability,binetteReiter2024estimands,pearlBareinboim2014transportability}. None by itself guarantees that the target of one analysis is the source of the next, which Section~\ref{ssec:neighbours} takes up.

Argument-based validity already has the resources to accommodate this principle. This paper's contribution is to make it usable for distributed AI evidence: typed descriptions of each study's source and target; five field comparisons that test whether purportedly adjacent links meet under the declared schema; two further requirements governing whether warrant transmits across the join; and a distinction among composition, convergence, and replacement. The principle identifies the failure mode; the audit supplies a procedure for diagnosing it.

Two obligations follow that fall outside any single link. A source report has to preserve the resolution a declared downstream claim will need, since an aggregate can leave several target-relevant states indistinguishable. And the evidence divides between developers and deployers, because no single party observes the whole chain.

Section~\ref{sec:projectibility} locates projectibility within validity theory and distinguishes it from construct meaning, prediction, and decision. Section~\ref{sec:composition} states the interface requirements and separates composition from convergence and replacement. Section~\ref{sec:legal} works through a legal-research case whose hypothetical results support one projection, defeat another, and leave a third unresolved. Section~\ref{sec:aggregation} shows how an upstream mean can erase the item-level differences a downstream projection needs. Sections~\ref{sec:agi} and \ref{sec:procedure} apply the result to general-capability claims and divide the evidential work between developers and deployers.

\section{Projectibility within a Validity Argument}\label{sec:projectibility}

\subsection{Rival extensions from the same observations}\label{ssec:goodman}

Goodman's new riddle of induction begins with observations that fit incompatible extensions equally well. Every emerald examined before time \(t\) is green. The observations fit the familiar hypothesis that emeralds are green. They also fit the rival hypothesis that emeralds are \mention{grue}. Under this hypothesis, the observed emeralds qualify because they're green, but an emerald first examined after \(t\) would qualify only if it were blue. The hypotheses agree on every observed emerald and disagree on the next unexamined one. Fit to the source observations doesn't determine which predicate is fit for projection \citep[74--80]{goodman1983}.

The benchmark analogue isn't that evaluators literally invent time-indexed predicates. Many classifications of unobserved cases agree on the scored items. Success can be grouped under \mention{general legal reasoning}, under \mention{answering supplied-text legal questions}, under \mention{recognizing patterns in questions drawn from familiar sources}, or under still narrower descriptions. Those predicates may be extensionally equivalent on the test set and diverge on a research request that requires current authority, live retrieval, source comparison, and a cited memo. More observations sampled under the same narrow item-generating rule may distinguish none of them.

\Textcite[84--98]{goodman1983} appeals to entrenchment, the history of successful use of a predicate in past projections. Other accounts place more weight on revisable background theory and causal structure \citep{boyd1991,khalidi2013}. This account needs no single sufficient criterion. Its more modest lesson is that the rule of extension has to be exposed. This doesn't remove induction; it makes a particular induction inspectable and revisable.

\subsection{Definition and scope}\label{ssec:definition}

A \term{projection} extends an interpretation, prediction, or explanation from specified source observations to unobserved cases. \term{Projectibility} concerns whether that bounded extension is warranted relative to a declared target and use. Three restrictions follow.

First, projectibility applies to a particular inference, not to a benchmark or system in the abstract. Its bearer is one bounded extension: this result, to these unobserved cases, under these conditions. The same benchmark result may provide substantial warrant for performance on further items generated under a registered template and little warrant for performance on open-ended professional work. Calling the benchmark \enquote{projectible} without naming the target suppresses the point at issue.

Second, support for a fixed projection is graded, but projectibility can't be summarized by a single scalar independent of the claim, scope, loss, and evidential standard. Support, scope, and the kinds of evidence bearing on the claim shouldn't be collapsed onto one scale. A report should preserve heterogeneous results: a predictive estimate and interval, a defeated invariance assumption, an unresolved population gap, and a supported narrow use. Collapsing these into a projectibility score recreates the aggregation problem. The status should remain claim-specific: supported under a stated evidential standard, defeated by specified evidence, or unresolved because relevant evidence is absent or imprecise. Each status should name its limit, such as support for one request type or dependence on evidence that only another party can supply.

Third, source fit is only part of the evidence. Warrant for a projection commonly comes from four places. Direct target sampling shows what happens in cases admitted by the target rule. Deliberate variation tests features suspected of changing the result while preserving what the claim treats as invariant. Background knowledge explains why some differences are plausible defeaters and others aren't. Replication across genuinely new levels (another template family, system lineage, site, or period) tests whether an earlier regularity survives the dimension being projected over. None can be replaced by a bare assertion that source and target are similar.

This account belongs inside an argument-based approach to validity. Kane represents a proposed interpretation and use as a sequence of inferences and assumptions, each requiring support commensurate with the ambition of the claim \citep[1--3, 22--25]{kane2013validating}. Messick's unified account likewise treats validity as an evidential judgment about score meaning and use, drawing on content, response processes, internal structure, relations with external variables, and consequences \citep[741--749]{messick1995Validity}. Projectibility concerns whether those parts of the argument that extend beyond cases observed under the source design are warranted.

Not every validity problem is a projection problem. A rubric that mis-scores the observed responses fails before any extension is attempted. A capability whose meaning is unclear poses an interpretive problem. Those failures may block a projection, but projectibility doesn't replace their specialized analysis.

\subsection{What projectibility adds to neighbouring frameworks}\label{ssec:neighbours}

Neighbouring work can be sorted by how far along the inference it reaches. Some frameworks stop at what a benchmark collects; others follow a single claim from evidence to interpretation and use. This paper asks what warrants the join between two such claims.

\Textcite{raji2021whole} argue that broad benchmark claims routinely outrun the contextual tasks used to construct them, and \Textcite{bowmanDahl2021benchmarking} set out how construct validity fails in language-model evaluation. \Textcite{liu2024ecbd} adapt evidence-centred design from educational assessment, formalizing benchmark construction into modules that require designers to describe, justify, and support each choice. Their subject is the evidence a benchmark collects about the capabilities it declares; movement beyond the benchmark isn't their topic.

Recent validity-centred work on AI reaches further along the inference. \Textcite{salaudeen2025measurement} offer a claim-aware framework that maps measurements and evaluations to claims, differentiates forms of validity, and recognizes that measurement producers and downstream claimants may be different stakeholders. \Textcite{freieslebenZezulka2025benchmarking} treat predictive benchmarks as measurement tools and specify internal, external, content, consequential, and auxiliary conditions for scientific inferences. \Textcite{bean2025measuring} document construct-validity weaknesses across 445 language-model benchmarks and give practical recommendations for benchmark development. This paper accepts the shared premise: the inferential object is a particular claim, and stronger claims need more evidence.

Claim-centred frameworks can show what evidence would support a benchmark interpretation and what would support a deployment criterion. A further question is whether the first conclusion actually supplies a premise for the second. A study of a base model answering supplied-text questions and a study of reviewed, retrieval-assisted memos may both concern legal work without forming a chain. Their evidence may compose, converge on a narrower claim, replace the need for one another for a particular purpose, or concern different objects altogether. The projectibility audit makes that handoff an explicit object of analysis.

\Textcite{freiesleben2026establishing} argues that the inferential account provides insufficient resources for articulating the meaning of theoretical capability constructs, and that capability benchmarks should instead be embedded in nomological networks. That objection concerns a use of argument-based validity not made here. This paper neither proposes a meaning for \mention{reasoning} nor infers the possession of reasoning from benchmark scores. It uses Kane's framework for the task it directly addresses: representing a proposed interpretation or use as a sequence of inferences and assumptions whose evidential support can be assessed. A substantive construct attribution, where one is proposed, needs whatever nomological, process, or causal support that attribution requires; the projectibility audit doesn't supply it.

Even a well-supported capability attribution doesn't carry a benchmark result into practice. A nomological network may show that task scores relate to a reasoning construct, or that the construct predicts an external criterion in a specified population. It still doesn't show that the relation survives when the base model becomes a retrieval-augmented application, that lawyers catch the application's errors, or that reviewed outputs produce the consequences assumed by policy. Each is a further projection over a different object, population, or condition. The empirical interface audit can therefore proceed whether the capability attribution is supported, rejected, or omitted.

Every study can estimate exactly what it says it estimates and the chain can still fail. One study may estimate mean benchmark performance, another the probability of a defect after lawyer review, and a third the rate at which defects reach clients. Each estimand may be precisely specified over its own population and data-acquisition rule, and \textcite{binetteReiter2024estimands} show how benchmark conclusions become confused when those elements are left implicit. The chain remains unsupported if the first quantity hasn't been shown to predict the second, or if the second was measured under conditions that don't match the third. Estimand specification identifies the quantities; the interface audit asks whether they can be joined.

Generalizability theory is the nearest measurement-side neighbour. A G-study estimates variance components over a declared \term{universe of admissible observations}, and a D-study generalizes to a universe that may be narrower; Brennan's own example has two investigators generalizing differently from the same components \citep[14]{brennan2001generalizability}. The framework therefore names the difference at issue here, within one design. The audit asks it across two: when separately conducted studies declare different universes, neither design contains the facet that would show whether they meet. Widening a G-study to span both is one legitimate form of the bridge that's owed.

Causal transportability supplies stronger formal results for a narrower kind of projection. Selection diagrams encode differences between source and target domains and establish when a causal relation is identifiable in the target from source experiments and target observations \citep{pearlBareinboim2014transportability}. Projectibility is broader: it covers descriptive generalization, construct interpretation, prediction, explanation, and sociotechnical outcomes, and it claims no identification theorem. Where the target claim is causal, a projectibility audit should defer to the corresponding causal requirements rather than treating generic robustness evidence as enough. Table~\ref{tab:frameworks} summarizes the division of labour; its final column states the question that motivates this paper.

The objection that the principle adds nothing to Kane deserves a direct answer. Kane treats validation as an evaluation of the \enquote{coherence and completeness} of an interpretation-and-use argument together with the plausibility of its inferences and assumptions \citep[1]{kane2013validating}. An argument specified to that standard can represent the interface assumptions among its inferences, and validating it would require supporting them. The principle stated here isn't a gap in argument-based validity.

Kane's own summary of chain reasoning shows what still has to be established: \enquote{A chain of reasoning is only as strong as its weakest link (Crooks et al., 1996), and strong evidence for part of an argument does not compensate for weaknesses in other parts of the argument} \citep[64]{kane2013validating}. That principle presupposes that the inferences form a chain. Whether they do is a prior question, and where the answer is no, the individually strong links don't warrant the endpoint by composition.

AI evidence is commonly produced in separate studies by different actors and assembled into a larger argument only later. The practical question is then whether their conclusions actually form a chain, which is a question about the joins rather than about any study. This paper specializes argument-based validity for that setting: it represents each endpoint as a typed record, compares purported joins field by field, and distinguishes links that never meet from links that meet but fail to transmit warrant. Kane's framework already has room for the assumptions at those interfaces; the audit makes their identification systematic.

\begin{table}[!t]
\centering\small
\caption{Neighbouring frameworks and the composition question}\label{tab:frameworks}
\begin{tabular}{@{}>{\raggedright\arraybackslash}p{0.20\linewidth}
                    >{\raggedright\arraybackslash}p{0.25\linewidth}
                    >{\raggedright\arraybackslash}p{0.27\linewidth}
                    >{\raggedright\arraybackslash}p{0.20\linewidth}@{}}
\toprule
Framework & Primary object & Question answered & Further composition question \\
\midrule
Argument-based validity & Interpretation-and-use argument & Which assumptions and evidence support a stated interpretation or use? & Does one supported conclusion supply the premise required by the next link? \\
Nomological network & Construct and its theoretical and empirical relations & What gives a capability construct content, and do its predicted relations hold? & Do those relations survive a change of task, system, site, or time? \\
Estimand specification & Metric, population, data acquisition, and aggregation & Which quantity does this analysis estimate? & Do adjacent links estimate quantities that can be composed, including any conditioning each requires? \\
Generalizability theory & A universe of admissible observations and its variance components & How far does a score generalize across the facets this design declared? & Do two studies declare the same universe, and if not, which facet bridges them? \\
Causal transportability & Causal relation across source and target domains & Under which assumptions is a target causal effect identifiable? & Are later interpretive, workflow, or decision links also supported? \\
Projectibility audit & Specified extension and its interfaces & What warrants this source--target extension, and can it be joined to adjacent extensions? & This is its subject: an explicit interface record and a diagnostic classification. \\
\bottomrule
\end{tabular}
\end{table}

\section{Why Warranted Links Don't Automatically Compose}\label{sec:composition}

\subsection{Typed nodes and projection edges}\label{ssec:typed}

The easiest way to conceal an inferential gap is to give both sides the same broad noun. A benchmark contains \mention{legal tasks}, and a firm performs \mention{legal tasks}; the shared label is taken to show that the score transfers. A developer tests \mention{the model}, and a deployer uses \mention{the model}; the shared label is taken to show continuity between the tested and deployed objects.

I use \term{type} here in the type-theoretic sense, specifically as a \term{record type}: roughly, a schema whose instances assign values to labelled fields \citep[sec.~11.8]{pierce2002types}. An instance of such a schema is a \term{node}. For this audit, an empirical node has five fields: the object evaluated, the population of cases, the conditions under which the evaluation occurs, the outcome recorded, and the period to which the claim applies. These types are revisable representations for an audit, not claims that the objects share an essence or form a natural kind.

A particular benchmark or deployment is represented as a node by filling those fields with values. Here, \mention{object} names a field, whereas a named model build is a possible value of that field. Likewise, \mention{legal tasks} may be intended as a value for the population field, but it supplies no case-generating or inclusion rule by which the benchmark and deployment populations could be shown to match.

Write an empirical node as
\begin{equation}\label{eq:node}
\mathcal{N}=\langle\text{object},\ \text{population},\ \text{conditions},\ \text{outcome},\ \text{period}\rangle.
\end{equation}
The population is a case-generating or inclusion rule, not just a list of cases, and the period covers the relevant system version as well as the span of time.

In a benchmark node, the object might be a named model build; the population, held-out items produced under specified templates; the conditions, the prompt, decoding procedure, tool access, and scorer; the outcome, item correctness; and the period, the evaluation date. In a deployment node, the object might instead be a model, retrieval index, prompt stack, and lawyer-review procedure; the population, a rule admitting two kinds of logged research request; the conditions, the firm's database and review instructions; the outcome, a substantive defect remaining in the final memo; and the period, the next six months.

Typing doesn't guarantee that two nodes match. It makes the proposed match inspectable. The benchmark and deployment above might both be described as involving \mention{legal tasks}, but their corresponding fields differ throughout. Repetition of the broad label supplies no warrant for the projection.

A capability attribution such as \mention{general reasoning} isn't another sampled endpoint. It's an interpretive claim about a bearer, supported where appropriate by content, process, causal, and nomological evidence. \Textcite[290]{cronbach1955construct} state the constraint on it exactly: a construct \enquote{is not \enquote{reduced} to the observations, but only combined with other constructs in the net to make predictions about observables}. It may be the terminal conclusion of one argument or a substantive premise in another.

What it doesn't carry is a sampled population of cases, a criterion outcome, or the conditions under which the attributed capability is expected to predict that outcome, which is what the next link would have to draw on. Where a capability attribution is invoked to license a step between empirical nodes, the audit asks which observable relation is required and whether that relation has been tested in the declared population.

Projections in the broad sense of Section~\ref{ssec:definition} include interpretive ones; the formalism here represents the empirical subset, whose endpoints can serve as interfaces between studies. An \term{edge} \(e\) is one such projection, running from a source node to a target node. An empirical projection edge includes its source and target nodes, projective claim \(C_e\), assumptions \(A_e\), and evidence \(E_e\) bearing on the claim and assumptions. Write \(e=\langle\mathcal{N}_s,\mathcal{N}_t,C_e,A_e,E_e\rangle\), also \(e:\mathcal{N}_s\rightsquigarrow\mathcal{N}_t\).

Generalizing from observed benchmark items to further items under the same generating rule is one empirical edge. Extrapolating to a new task population, transporting a relation to a new model build, and predicting a post-review outcome are further edges. Interpretive inferences remain part of the broader validity argument, but they aren't themselves sample-bearing interfaces between empirical studies. The notation forces no commitment that every argument has the same sequence. It prevents an argument from treating a sequence as one undifferentiated leap.

\begin{figure}[!t]
\centering
\footnotesize
\textbf{A. Claimed benchmark-to-use chain}\par\vspace{3.5mm}
\resizebox{\linewidth}{!}{%
\begin{tikzpicture}[
  node distance=5mm,
  box/.style={draw, rounded corners=2pt, align=center, inner sep=4pt,
              text width=21mm, minimum height=14mm, font=\footnotesize},
  cbox/.style={draw, dashed, rounded corners=2pt, align=center, inner sep=4pt,
              text width=24mm, minimum height=12mm, font=\footnotesize},
  flow/.style={-{Latex[length=2mm]}, thick},
  side/.style={-{Latex[length=2mm]}, dashed, semithick},
  broken/.style={dashed, semithick},
  lab/.style={align=center, font=\scriptsize, text width=29mm},
  elab/.style={align=center, font=\scriptsize, text width=25mm},
  slab/.style={align=center, font=\scriptsize, text width=28mm}
]
\node[box] (obs) {Observed benchmark\\responses\\[1pt]$\mathcal{N}_0$};
\node[box, right=8mm of obs] (uni) {Further test-rule\\performance\\[1pt]$\mathcal{N}_1/\mathcal{N}'_1$};
\node[box, right=8mm of uni] (app) {Application outputs on\\eligible target tasks\\[1pt]$\mathcal{N}_2/\mathcal{N}'_2$};
\node[box, right=8mm of app] (rev) {Final work after\\registered review\\[1pt]$\mathcal{N}_3/\mathcal{N}'_3$};
\node[box, right=8mm of rev] (out) {Consequences under\\authorized policy\\[1pt]$\mathcal{N}_4$};

\path (uni.north) -- (app.north) coordinate[midway] (capanchor);
\node[cbox, above=15mm of capanchor] (cap) {Capability\\attribution};

\draw[flow] (obs) -- node[lab, above=10mm] {$e_1$ generalization}
                     node[elab, below=10mm] {template-family holdout} (uni);
\draw[flow] (uni) -- node[lab, above=10mm] {$e_2$ task/system projection}
                     node[elab, below=10mm] {target-task and configuration sample} (app);
\draw[flow] (app) -- node[lab, above=10mm] {$e_3$ review projection}
                     node[elab, below=10mm] {paired draft and final scoring} (rev);
\draw[flow] (rev) -- node[lab, above=10mm] {$e_4$ outcome projection}
                     node[elab, below=10mm] {exposure and outcome surveillance} (out);

\draw[side] (uni.north) -- node[slab, pos=0.6, anchor=east, xshift=-1mm, text width=20mm, align=right] {interpretation} (cap.south west);
\path (app.north) ++(0,5mm) coordinate (gap);
\draw[broken] (cap.south east) -- node[slab, pos=0.5, anchor=west, xshift=1mm, text width=19mm, align=left] {criterion relation required} (gap);

\end{tikzpicture}%
}
\medskip
\textbf{B. Audit of separately warranted \(e_1\) and \(e_2\)}\par\smallskip
\resizebox{\linewidth}{!}{%
\begin{tikzpicture}[
  q/.style={draw, rounded corners=2pt, align=center, inner sep=4pt,
            text width=39mm, minimum height=20mm, font=\footnotesize},
  res/.style={draw, rounded corners=2pt, align=center, inner sep=4pt,
              text width=34mm, minimum height=11mm, font=\footnotesize},
  ok/.style={res, thick},
  fl/.style={-{Latex[length=2mm]}, thick},
  t/.style={font=\scriptsize, inner sep=1.5pt}
]
\node[q] (q1) {\textbf{1. Endpoints}\\[2pt]
  Do \(\mathcal{N}_1\), where \(e_1\) ends, and \(\mathcal{N}'_1\), where \(e_2\)'s own study begins, align in object, population, conditions, outcome, and period, or is the difference covered by a warranted bridge?};
\node[q, right=16mm of q1] (q2) {\textbf{2. Transmission}\\[2pt]
  Are assumptions and effect modifiers compatible, and is dependence carried through with the uncertainty?};
\node[ok, right=16mm of q2] (yes) {Composition warranted\\for the declared use};
\node[res, below=10mm of q1] (f1) {No warranted bridge:\\ill-typed as stated};
\node[res, below=10mm of q2] (f2) {Well-typed;\\warrant doesn't transmit};

\draw[fl] (q1) -- node[t, above] {yes} (q2);
\draw[fl] (q2) -- node[t, above] {yes} (yes);
\draw[fl] (q1) -- node[t, right] {no} (f1);
\draw[fl] (q2) -- node[t, right] {no} (f2);
\end{tikzpicture}%
}
\caption{A claimed benchmark-to-use chain and its interface audit. Panel A names the empirical nodes and projective edges. Each intermediate box carries two records drawn as one, the upstream target and the downstream source, written \(\mathcal{N}_i/\mathcal{N}'_i\); whether they are one is what panel B asks, and the question arises at every join. A capability attribution, supported where appropriate by content, process, causal, and nomological evidence, may serve as an interpretive premise for \(e_2\) but isn't an empirical endpoint. Panel B audits endpoint alignment before warrant transmission and shows the three possible outcomes. \(\mathcal{N}_0\) and \(\mathcal{N}_1\) describe the same object over different populations}\label{fig:chain}
\end{figure}
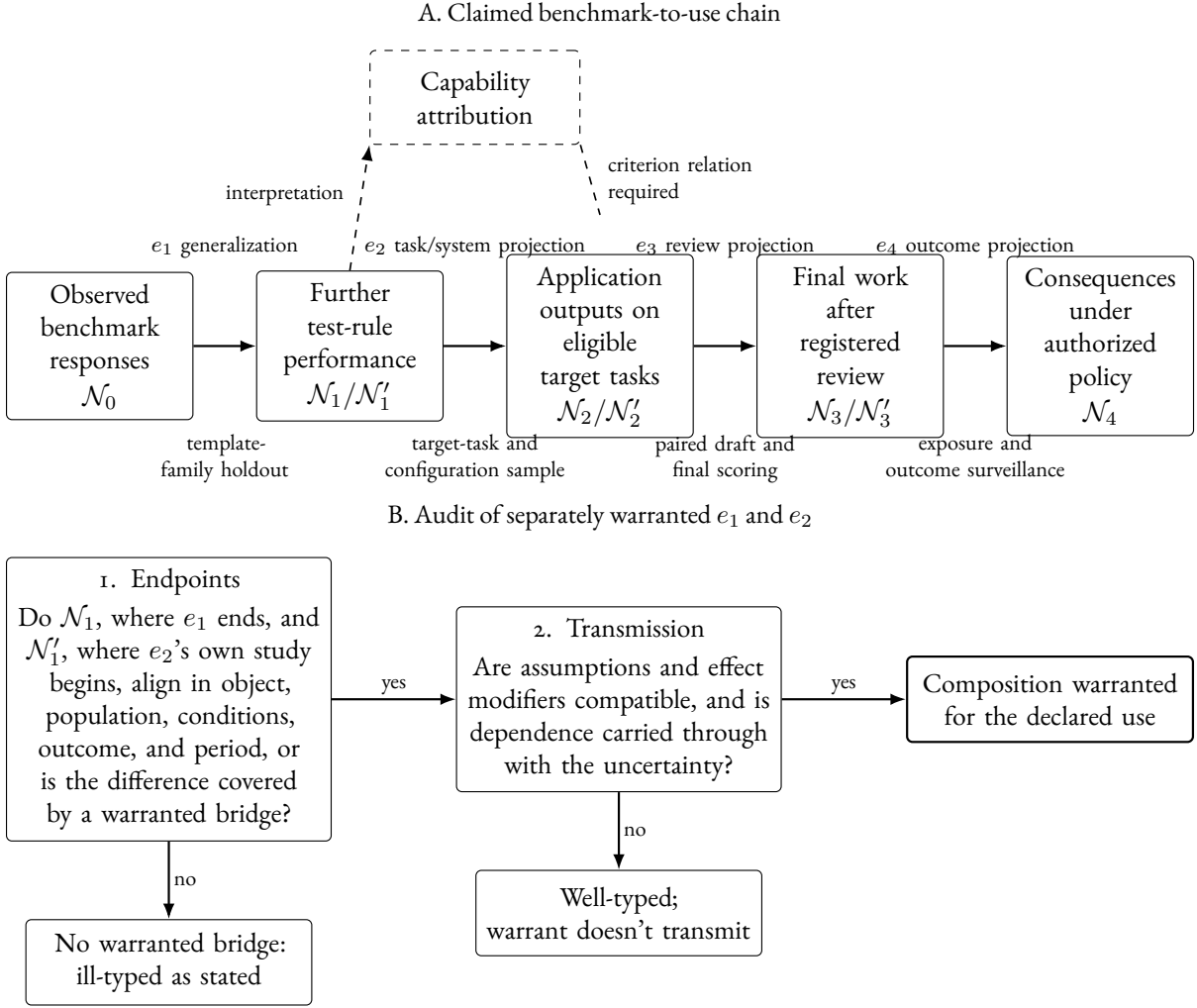

Fig.~\ref{fig:chain} names the nodes and edges and matches each projection to evidence that bears on it. No amount of evidence under one arrow automatically fills another.

\subsection{Endpoint alignment and warrant transmission}\label{ssec:interfaces}

Suppose \(e_1:\mathcal{N}_0\rightsquigarrow\mathcal{N}_1\) and \(e_2:\mathcal{N}_1'\rightsquigarrow\mathcal{N}_2\) are each supported. Two questions arise in order. First, do the links meet? This is an endpoint-alignment question about whether the proposed composition is well formed. Alignment requires continuity in object, population, conditions, outcome, and period, or a separately warranted bridge across the difference. Second, if they meet, does warrant transmit across the join? That further requires compatibility of assumptions and effect modifiers, and propagation of evidential dependence and uncertainty. The first failure leaves the proposed composition ill-typed as stated; the second leaves a well-typed composition unwarranted.

Take a concrete pair. One study finds that systems with higher benchmark scores produce fewer defective drafts; another finds that lawyer review removes a certain proportion of draft defects. It's tempting to combine the results to predict the quality of final work. But that combination makes two claims that neither study establishes alone: that both results apply to the same target population, and that review works in the same way across systems once draft quality is held fixed.

The five alignment requirements read off Equation~\ref{eq:node}; the two transmission requirements read off the assumptions \(A_e\) and the evidence \(E_e\) carried by the edges. All seven presuppose empirical nodes at both endpoints, which the previous subsection secured.

\term{Endpoint alignment} has five requirements. Where one fails, the mismatched field identifies the bridge that's owed. A result about supplied-text questions doesn't arrive at a population of live-retrieval requests.

\begin{enumerate}
\item Object continuity requires a declared system or workflow projection when the object changes. A base model, a model connected to a database, a complete application with citation validation, and a memo after lawyer review are different objects. Product names and version families don't establish continuity. Shared training, distillation, retrieval components, or evaluation data may create dependence without preserving behaviour.

\item Population alignment requires the cases reached by the upstream projection to have the distribution, support, and grouping assumed downstream. A study on short, self-contained questions doesn't supply cases for a study whose source consists of open-ended requests with disputed authorities. Even within one task label, a downstream study may sample routine files while the upstream claim includes novel, urgent, or multilingual work. The inclusion rule, not the label, determines alignment.

\item Condition alignment requires the operating conditions of the shared endpoint to match on both sides, or their difference to be shown harmless for the target claim. Two studies conducted under incompatible prompts, decoding settings, databases, retrieval configurations, or reviewer instructions don't compose merely because both are described as evaluating the same system.

\item Outcome and scale continuity requires the first edge to deliver the quantity the next uses. Benchmark accuracy, a factor score, the probability of a citation defect in a draft, a defect after review, and a client loss aren't interchangeable outcomes. Putting each measure on a zero-to-one scale changes its numerical range, not what a difference means. A change of .10 in benchmark accuracy isn't equivalent to a .10 change in defect probability or client loss. If the outcome changes, the relation between the measures is itself an inferential link.

\item Temporal alignment treats the period as a field like any other. A result established on one model build, database index, and work period doesn't reach a later one because the product name is unchanged. Where the downstream claim covers a span the upstream study didn't sample, the difference is a further projection, answered by a chronological holdout or a declared retest trigger rather than by an assumption of continuity.
\end{enumerate}

\term{Warrant transmission} has two further requirements once the endpoints align.

\begin{enumerate}
\item Assumptions and effect modifiers have to remain compatible across the join. Assumptions that support one edge can fail under the conditions of the next. Review may reduce one error class while increasing delay or inducing overreliance; retrieval may improve source coverage while introducing irrelevant but persuasive cases. Relevant \term{effect modifiers}, variables across whose strata the effect of interest differs, have to be carried to the interface rather than averaged away. They enter as conditioning variables rather than as a second intervention \citep[864]{vanderweele2009interaction}, which is what the composed prediction below turns on. An effect modifier dropped at the join can leave both component estimates correct and the composed prediction wrong.

\item Dependence and uncertainty have to be propagated rather than reset at the next edge. A chain can look better supported than it is when its links reuse the same items, scorer, model lineage, or development decisions. Twenty outputs from one build aren't twenty systems, and thirty-two comparisons among related models and shared benchmarks aren't thirty-two independent replications. Where estimates are composed, their covariance, selection, and uncertainty should be represented. Where a qualitative assumption is uncertain, the chain inherits that uncertainty.
\end{enumerate}

Return to the two studies above, one relating benchmark scores to draft defects and the other measuring what review removes. Let \(B\) be a benchmark result, \(D\) a draft-level outcome, and \(Z\) a final outcome after review. The probabilities below concern a system--request pair drawn at random from a target population \(T\): \(B\) varies across systems, while \(D\) and \(Z\) vary across pairs. In \(T\),
\begin{equation}\label{eq:composition_identity}
P_T(Z\mid B)
  =\sum_d P_T(Z\mid D=d,B)\,P_T(D=d\mid B).
\end{equation}
In words, consider each possible draft outcome. Multiply its probability at a given benchmark result by the probability of the final outcome after review for that draft and benchmark result, then add across the possible draft outcomes.

The first study may estimate \(P_T(D\mid B)\), while the second estimates \(P_T(Z\mid D)\). Equation~\ref{eq:composition_identity}, though, requires \(P_T(Z\mid D,B)\), the probability of the final outcome given the draft outcome and the benchmark result together. Substituting the second quantity is warranted if the benchmark result adds no information about review success once the draft outcome is known. That relation is conditional independence, written \(Z\perp B\mid D\). Short of that, the substitution rests on the discrepancies happening to cancel in the weighted sum, which is not something an evaluator is usually in a position to claim.

That independence requirement could fail if, for example, high-scoring systems produce more fluent errors that reviewers are likelier to trust. And if either study concerns a different population, it doesn't supply the corresponding relation in \(T\). Perfectly aligned endpoint descriptions still don't ensure that warrant passes across the join. Formal causal claims require more, but the problem arises even for descriptive prediction.

The principle may look trivial, on the ground that warrant rarely transfers automatically and that the requirements above amount to a demand that two studies match. Table~\ref{tab:frameworks} answers the first: neighbouring frameworks can represent the assumptions an interface requires, while this audit makes the interface itself an explicit and reusable object of declaration and diagnosis. For the second, Equation~\ref{eq:composition_identity} displays the quantity a composition needs and the condition licensing the available substitute, and Section~\ref{ssec:counterexample} shows that condition failing while every field aligns.

\subsection{Composition, convergence, and replacement}\label{ssec:relations}

Three relations among studies are easily conflated. In \term{composition}, the conclusion of one study supplies a premise or population for the next. A benchmark score predicts draft defects across systems; draft defects predict post-review defects under a specified procedure. Those links compose only if their object, population, conditions, outcome, and period align at the shared endpoint, and only if warrant then transmits across it.

In \term{convergence}, distinct evidence bears on the same claim without forming a chain. A benchmark, expert assessment, and process intervention might each support a reasoning attribution. Their agreement can strengthen the claim, but no result is passed as an input from one study to another. Dependence still matters: three tests built from the same source materials don't provide three independent routes.

\Textcite{jung2026psychometric} supply a measured instance. Human psychometric instruments applied to language models reproduced the inter-test correlations that the theory behind those constructs predicts: sexism with racism at \(r_s=.47\), authority with benevolent sexism at \(.43\), and fairness with hostile sexism at \(-.37\). Those agreeing measurements bear on what the instruments measure. They don't bear on how the models then behaved in matched downstream settings, which Section~\ref{ssec:chc_roles} takes up.

In \term{replacement}, direct evidence from the target setting bypasses the ambitious upstream projection rather than extending it. A firm can sample its own research requests and measure defects in the final, lawyer-reviewed memos. That local study can support a policy for this workflow without first showing that the base model possesses general legal-reasoning ability. It doesn't validate the benchmark's broader claim; it begins a new, narrower inference from observations closer to the intended use. This is often epistemically preferable: the workflow can be supported while the broader capability question remains open.

The distinctions matter because the phrase \mention{benchmark plus local validation} can describe all three. If the local study compares several systems and tests whether their benchmark \term{profiles} (the vectors of domain scores a battery reports) predict their defect rates on the same requests, it tests a composable predictive edge. If it evaluates only one frozen application, the benchmark profile is constant across its requests and can't explain which request fails. The local study then replaces, rather than confirms, the benchmark-to-use projection. If both studies are cited merely because they sound favourable, they provide neither composition nor well-characterized convergence.

\subsection{Two counterexamples}\label{ssec:counterexample}

The two failures separated above look alike on the page and come apart under inspection. The first is \term{spurious adjacency}, where the links never meet. Consider two individually true claims. First, a model answers 90\% of held-out supplied-text legal questions correctly under the benchmark protocol. Second, lawyers following a written review instruction catch 99\% of fabricated citations in drafts that a retrieval-based application produces for a set of routine research requests. Neither result is defective on its own. Their conjunction still doesn't warrant that reviewed memos are substantively correct.

Neither study measures the failure that matters: the application's retrieval may omit controlling authority without fabricating any citation. The benchmark never tests retrieval, and the review study counts only fabricated citations. Suppose the application omits a controlling case on 20\% of requests, the omission isn't visible from the citations that remain, and the lawyers' procedure doesn't require an independent update search. Both premises remain true while one fifth of final memos contain a substantive defect. The chain fails at two interfaces: the outcome changes from supplied-text correctness to citation fabrication, and the review evidence doesn't cover omitted authority. Multiplying the two percentages obscures rather than repairs the gap.

The counterexample also shows what positive evidence would look like. The firm could prepare an authority list independently of the system, score omitted controlling authority as well as citation fabrication, and require reviewers to compare every final memo with that list or conduct a defined update search. The revised study wouldn't make the benchmark general. It would supply observations for the missing workflow link.

Something close to this has already happened. Legal-research vendors advertised retrieval-grounded products as delivering \enquote{100\% hallucination-free linked legal citations} or as avoiding hallucination by relying on trusted content, and \textcite[234]{magesh2025legalhallucination} found that no empirical evidence accompanied those claims. Their preregistered evaluation reports hallucination on more than 17\% of queries for both the LexisNexis and Thomson Reuters tools, and incomplete answers on more than 60\% for Thomson Reuters. The advertised inference runs from a component property, retrieval over an authoritative database, to what the product asserts, with no observations at the join. Incompleteness is the omission failure isolated above rather than the fabrication failure the marketing addressed.

The second, \term{transmission failure}, is harder to see, because the links do meet. Suppose several systems are evaluated on the same held-out requests, and a benchmark score predicts the frequency of draft defects across them. Suppose reviewers working from one instruction catch 90\% of draft defects across drafts produced by all of those systems on those same requests. The draft-defect variable is now the same variable in both studies, over the same systems and the same requests, so every field of the shared endpoint matches and no bridge is owed. The composed prediction can still be wrong. If high-scoring systems fail mostly by omitting an authority that no citation flags, while low-scoring systems fail mostly by inventing a citation that any reviewer catches, then review effectiveness depends on which system produced the draft. In the terms of Equation~\ref{eq:composition_identity}, \(P_T(Z\mid D,B)\neq P_T(Z\mid D)\), and a catch rate averaged over the whole set overstates what review achieves for exactly the systems the benchmark ranks highest. Both component relations are well estimated, the interface is aligned, and the chain still fails. Nothing here is repaired by checking that the two studies denote the same things; the defect is an effect modifier dropped at the join.

A natural objection collapses the second failure into the first. If review effectiveness depends on defect type, then \mention{draft defect} was the wrong outcome to declare, and a finer type would have exposed a mismatch. The objection is right that the two failures are distinguished relative to a declared schema. Representing the shared endpoint as a pair, or adding a field for conditioning structure, would make this case a mismatch.

That relocation is a repair, and an audit should permit it. The five fields are a minimum schema, refinable when a proposed join needs more resolution, and what refinement achieves is to move a failure from the second stage to the first, where comparing declared records can catch it. Alignment stays a relation among declared fields, transmission a relation among the assumptions and evidence the edges carry.

Spurious adjacency is primarily a record-comparison problem; transmission failure is a substantive modelling problem. Comparing the declared endpoint descriptions can expose the first without any domain theory. The second remains after those descriptions match, and finding it requires knowledge of effect modifiers, dependence, or conditioning structure. The repairs differ too. Spurious adjacency normally calls for an explicit bridge or new observations at the missing endpoint. Transmission failure can sometimes be repaired by reanalysing data already collected, but only where the source report preserved the distinction the composed inference needs, which is the information requirement Section~\ref{sec:aggregation} takes up.

\section{A Worked Projectibility Audit}\label{sec:legal}

\subsection{The source result and the proposed use}\label{ssec:legal_setup}

Suppose a developer reports a high legal-reasoning score for a base model. The battery presents each question with a fixed set of materials and marks the answer against a reference response. Its held-out design may support generalization to further questions produced under the same item rules. The score doesn't observe whether the model can find current law, distinguish controlling from merely similar authority, produce a linked citation, or work within a lawyer's review procedure.

An Ontario employment-law group is considering a product built from the model. The application receives a lawyer's research request, searches the firm's licensed legal database, and drafts a one-page memo. The firm proposes to authorize it only for two kinds of request: whether a termination clause is enforceable and what period of reasonable notice is likely. An eligible request is written in English, concerns Ontario law, identifies the relevant dates and employment facts, and asks one of those two questions. Requests concerning another jurisdiction, a different subject, missing facts that prevent research, or a novel constitutional issue are outside the target.

Each memo must identify the issue, state the governing propositions, cite the current controlling authorities, link each proposition to a passage that supports it, and undergo line-by-line lawyer review before it leaves the firm.

The object under consideration isn't the benchmarked base model. It's a frozen configuration consisting of the model build, legal-database index, retrieval settings, system prompt, drafting template, citation-validation component, and review instruction. The final outcome belongs to the complete workflow, not to the model alone. That distinction is familiar in sociotechnical analyses of AI: component accuracy doesn't by itself determine the quality of decisions or work produced after human interaction \citep{dobbeWolters2024sociotechnical,buijsman2026accuracy,langer2025oversight}.

The rates \textcite{magesh2025legalhallucination} measured for commercial legal-research tools don't estimate the defect rate of this application. They do identify a plausible defeater and an outcome the firm should record.

\subsection{The declared chain}\label{ssec:legal_chain}

Table~\ref{tab:legal_chain} states the proposed links before any local result is known. The source and target columns are deliberately repetitive: they show where an apparently continuous claim changes object, population, or outcome.

\begin{table}[!t]
\centering\small
\caption{The legal-assistant argument decomposed into projection links}\label{tab:legal_chain}
\begin{tabular}{@{}>{\raggedright\arraybackslash}p{0.05\linewidth}
                    >{\raggedright\arraybackslash}p{0.23\linewidth}
                    >{\raggedright\arraybackslash}p{0.25\linewidth}
                    >{\raggedright\arraybackslash}p{0.38\linewidth}@{}}
\toprule
 & Source & Target & Evidence that bears on the link \\
\midrule
1 & Scored supplied-text questions & Further questions under the registered benchmark rules & Hold out whole templates or item families; document contamination checks and scorer reliability \\
2 & Performance under the benchmark rules & Draft memos for eligible local requests & Sample logged requests by the inclusion rule; prepare authority lists independently; score draft defects \\
3 & Frozen application drafts & Final memos after the written review procedure & Score the same outputs before and after review; record who reviewed them, what was corrected, and how long review took \\
4 & Ordinary retrieval condition & Retrieval with topically similar but legally irrelevant cases & Use an answer-preserving alteration not seen during prompt or rubric development; keep it out of tuning \\
5 & Termination-clause requests & Reasonable-notice requests & Sample and report each request type separately; don't infer one from the other by the label \mention{employment law} \\
6 & Current build and database index & Later build, index, or six-month work period & Use a chronological holdout and retest after a material component change \\
7 & Estimated defects and review costs & Authorization of a policy & Compare no assistant, search suggestions, mandatory-review drafting, and broader use under stated values and constraints \\
\bottomrule
\end{tabular}
\end{table}

Link~2 isn't established by showing that the battery contains several legal domains. It requires observations from the firm's task. Link~3 isn't established by assigning a lawyer to each output. Review effectiveness is itself an empirical relation, and it requires scoring the same outputs before and after review. Link~5 isn't established by averaging both request types; it requires a separate estimate for each. Link~7 isn't a projection from cases alone. It combines empirical premises with values, feasible alternatives, professional obligations, and institutional authority.

\subsection{A confirmatory local design}\label{ssec:legal_design}

A design that supplies those observations has to keep development separate from confirmation, so that the evidence testing a claim isn't the evidence the claim was built on. The one below is a deliberately demanding illustration of what a broad, quantified authorization could require, not a minimum; Section~\ref{sec:limitations} considers lighter evidence for narrower claims. It supplies direct evidence for links~2--5; it doesn't test link~6 unless the later period or changed component is sampled, and it doesn't settle link~7.

The firm develops its prompt, inclusion rule, reference-answer procedure, and scoring rubric on older closed files. Every request from one client file remains in the same split. Confirmation uses a later untouched block containing 200 eligible termination-clause requests and 200 eligible reasonable-notice requests, with one request selected from each client file. The counts are illustrative, but each number in the design has a function: 200 observations per declared stratum allow the firm to assess its chosen defect tolerance separately rather than letting the more frequent or easier request type dominate a pooled estimate.

Before any output is generated, two lawyers who supervise these file types independently prepare, for each request, a list of controlling authorities and the legal propositions for which each authority is required. They resolve disagreements before seeing the application's memo, and their initial disagreement is recorded rather than discarded, with cases they can't settle marked as disputed instead of forced to consensus. The procedure avoids defining the reference standard around whatever the system happened to retrieve, though it doesn't make the standard observation-free.

For each request, the application produces one draft under the ordinary retrieval setting. It also produces a second draft after topically similar but legally irrelevant cases are inserted near the top of the retrieved set. The alteration leaves the correct legal answer unchanged and tests whether retrieval rank rather than legal relevance controls the memo. The two drafts from one request are assigned to different reviewing lawyers, and no lawyer sees both versions. Order is counterbalanced. Blind scorers assess each draft, the assigned lawyers conduct the registered review, and different blind scorers assess the final memos.

A \term{substantive defect} is present when at least one citation doesn't resolve, a cited passage doesn't support the proposition attached to it, or the memo omits an authority on the independently prepared controlling-authority list. The three defect types are also reported separately. An unresolvable citation remaining in a final memo is a red-line outcome; it can't be offset by correct propositions elsewhere. Review time is recorded in minutes and kept out of the defect measure, since minutes and defects aren't on a common scale. It enters the policy comparison as minutes.

The firm sets the confirmatory rule before opening the later block. For each request type and condition, support for the empirical performance premise of mandatory-review use requires both (i) a one-sided 95\% exact upper confidence bound below a post-review substantive-defect tolerance of 3\% and (ii) no unresolvable citation in a final memo. The claim is defeated if the corresponding lower bound exceeds 3\% or if a red-line citation remains. Other results are unresolved.

Neither the 3\% tolerance nor the 95\% level is supplied by the benchmark or proposed as a general legal standard. Both are the firm's to justify against the consequences and costs of the available policies: the tolerance fixes how much residual defect the firm will accept, the level how much evidence it wants before saying so.

The three labels summarize the registered rule; they don't replace the estimate. The red line especially is a decision constraint rather than a measurement, since observing no unresolvable citation in 200 memos doesn't establish that the rate is zero. Whichever way the rule falls, the report carries the observed rates and the relevant bounds.

Four request-type and condition combinations are tested and authorization is sought only for those that pass, so the family of candidate authorizations is registered in advance and the criterion applied to each is stated as one of four rather than as a single test. The four claims are interpreted separately, and the procedure makes no family-level false-confirmation guarantee. A firm that needs one, because any passing condition will be selected for authorization, should register simultaneous bounds or a hierarchical analysis instead.

Fixing the rule in advance narrows analyst discretion without eliminating it. Adjudicating a borderline citation, pooling the two request types or reporting them apart, and handling a request that yields two outputs all remain open, and each can move the estimate. The audit records those choices and treats unplanned alternatives as exploratory rather than as independent confirmations \citep{GelmanLoken2013}.

The exact bounds keep the constructed example transparent by treating the file-level outcomes as exchangeable Bernoulli draws, at the cost of being conservative: the altered termination-clause row is unresolved on an upper bound of 3.11\% against a 3\% tolerance, a margin that reflects the interval's conservatism as much as the evidence. An actual study in which the same lawyers review many files should register a clustered or hierarchical uncertainty analysis and apply its support criterion to that estimate. The inferential point doesn't depend on the simple bound.

The design doesn't test whether the developer's benchmark score predicts local defects. With one frozen application, the benchmark profile is constant across the 400 requests. A request-level prediction requires variables that vary by request, such as request type, the number and age of retrieved authorities, conflicting appellate treatment, or exposure to the registered alteration. A benchmark-to-local predictive study would instead need several genuinely distinct systems evaluated on the same held-out requests.

\subsection{Hypothetical results and their inferential status}\label{ssec:legal_results}

Table~\ref{tab:hypothetical_results} gives constructed results. They aren't observations about any commercial system. Their purpose is to show that the same audit can support, defeat, and suspend judgment about different projections without converting those statuses into one overall verdict.

\begin{table}[!t]
\centering\small
\setlength{\tabcolsep}{3pt}
\caption{Constructed confirmatory results for the worked example. Exact bounds are one-sided 95\% bounds for the post-review defect probability}\label{tab:hypothetical_results}
\begin{tabular}{@{}>{\raggedright\arraybackslash}p{0.23\linewidth}
                    >{\centering\arraybackslash}p{0.13\linewidth}
                    >{\centering\arraybackslash}p{0.14\linewidth}
                    >{\centering\arraybackslash}p{0.16\linewidth}
                    >{\centering\arraybackslash}p{0.12\linewidth}
                    >{\raggedright\arraybackslash}p{0.13\linewidth}@{}}
\toprule
Request type and condition & Draft defects & Final defects & Relevant exact bound & Final red-line citations & Status \\
\midrule
Termination clause, ordinary retrieval & 38/200 & 1/200 & upper 2.35\% & 0 & Supported \\
Termination clause, irrelevant-case alteration & 46/200 & 2/200 & upper 3.11\% & 0 & Unresolved \\
Reasonable notice, ordinary retrieval & 52/200 & 18/200 & lower 5.90\% & 2 & Defeated \\
Reasonable notice, irrelevant-case alteration & 61/200 & 24/200 & lower 8.41\% & 3 & Defeated \\
Later database index or model build & -- & -- & no estimate & not observed & Unresolved \\
\bottomrule
\end{tabular}
\end{table}

The ordinary termination-clause claim is supported under the registered rule. The upper bound is below the firm's tolerance and no red-line citation survives. The result warrants a narrow projection from the sampled later files to other requests admitted by the same rule under the same frozen configuration and review procedure. It doesn't warrant unreviewed drafting, another office, another request type, or a later build.

The termination-clause alteration result is unresolved, not a failure disguised as success and not evidence of robustness. Under the alteration the reasonable-notice claim is defeated as it already was, so the altered condition adds nothing there. Its point estimate is low, but the registered upper bound remains above 3\%. More importantly, the unresolved status identifies the exact missing precision. The firm can collect new alteration cases if the prospective value of that evidence justifies the cost; it can't pool the altered and ordinary outputs to make the interval narrower for a claim about the altered condition.

The reasonable-notice claim is defeated. Its lower bound exceeds the tolerance, and two unresolvable citations remain after review. Success on termination-clause requests can't compensate for that result. The failed claim can lead to a prespecified narrower policy (for example, search suggestions only on reasonable-notice files) but not to a post hoc target defined around the requests that happened to pass.

The future-build claim remains unresolved because no future build or database index appears in the observations. A chronological holdout from the current configuration doesn't become evidence about a materially changed configuration merely because the product name remains the same. Authorization can be conditioned on freezing the components or on retesting after a declared update trigger.

Table~\ref{tab:audit_status} returns these results to the complete chain. The crucial row is the benchmark-to-local-defects projection, link~2 in Table~\ref{tab:legal_chain}. The local draft study estimates the frozen application's performance on local requests, but it doesn't establish that the developer's benchmark score predicted that performance. For the firm's bounded decision, the direct local evidence replaces the benchmark-to-use projection. The benchmark remains useful as a description of the source evaluation and perhaps as a screening instrument for choosing systems to test. It doesn't acquire external reach from the local result.

\begin{table}[tbp]
\centering\small
\caption{Warrant status after the constructed local study}\label{tab:audit_status}
\begin{tabular}{@{}>{\raggedright\arraybackslash}p{0.30\linewidth}
                    >{\raggedright\arraybackslash}p{0.19\linewidth}
                    >{\raggedright\arraybackslash}p{0.42\linewidth}@{}}
\toprule
Projection & Status & Reason \\
\midrule
Observed benchmark items to further items under the benchmark rules & Conditional & Requires the developer's template-level holdout, contamination checks, and scorer evidence; the local study adds nothing to this link \\
Benchmark score to defects on local research drafts & Unresolved & One system supplies no across-system benchmark--criterion relation; the local request sample directly estimates draft defects instead \\
Frozen application to reviewed termination-clause memos under ordinary retrieval & Supported narrowly & The direct target sample meets the registered defect and red-line criteria under the specified review procedure \\
Ordinary to altered retrieval on termination-clause requests & Unresolved & The altered-condition upper bound misses the registered tolerance; the reasonable-notice claim is already defeated under ordinary retrieval \\
Termination-clause to reasonable-notice use & Defeated & The second request type has a separately observed post-review defect rate above tolerance and surviving red-line errors \\
Current configuration to a later index or build & Unresolved & The changed object and period weren't sampled \\
Empirical results to authorization & Not settled by projectibility alone & Authorization requires comparison of feasible policies, review time, confidentiality, professional obligations, and consequences \\
\bottomrule
\end{tabular}
\end{table}

\subsection{From empirical warrant to policy}\label{ssec:legal_policy}

Even the supported termination-clause result is only one premise in a decision. The firm has at least four feasible policies: no assistant, assistant-generated search suggestions, drafting with the registered line-by-line review, and drafting with a lighter check. It can compare them on final defects, unresolvable citations, omitted authorities, lawyer minutes, confidentiality exposure, and whether an error reaches a client-facing memo or filed document. A policy that reduces draft time while increasing final defects may be inferior; a policy with slightly more review time may be preferable if it eliminates red-line outcomes.

This separation matters because human review isn't a fixed safety multiplier. Studies of human oversight show that detection depends on base rates, signal quality, reviewer incentives, workload, and the information supplied with an output \citep{langer2025oversight}. More accurate component systems can also produce worse joint performance when people under- or over-rely on them \citep{buijsman2026accuracy}. The legal study records the workflow's final result rather than multiplying a model accuracy by an assumed review rate.

Values enter before the final authorization as well as at it. Treating an unresolvable citation as a noncompensatory red line is a choice about what errors may trade off against other gains. Choosing 3\% as a tolerance and deciding whether lawyer minutes justify a narrower use are further choices. Validity evidence can show which empirical premises are supported; it can't choose the firm's loss function or discharge its professional duties \citep[747--749]{messick1995Validity}. Representing the authorization itself, rather than the premises feeding it, is a separate problem: it needs a semantics for what was delegated, over which operations, and for how long \ifanon\citep{anonSelfDelegationAssurance}\else\citep{reynolds2026delegationassurance}\fi.

\section{Aggregation Can Destroy Interface Evidence}\label{sec:aggregation}

A chain can fail even when the system and setting stay fixed, because the source report has discarded information the next study needs. A total score may answer the developer's descriptive question while making the deployer's target question unanswerable. This section separates net change, item-level movement, concentrated deterioration, and persistent absolute loss, reports what one released design looks like, and settles under a known truth whether the estimands can be told apart. The statistical quantities aren't offered as a universal scorecard. They show that what a source report must preserve depends on the downstream claim: aggregation is harmless only when it doesn't erase distinctions needed at the next interface.

\subsection{One stable mean, several incompatible states}\label{ssec:aggregate_logic}

Consider the same \(N\) items scored under a baseline condition \(0\) and an altered condition \(p\). Let \(s_{ic}\in[0,1]\) be the score for item \(i\) under condition \(c\), and let \(\delta_i=s_{ip}-s_{i0}\). Define the signed level change \(L\) as the mean item change:
\begin{equation}\label{eq:signed_change}
L=\frac{1}{N}\sum_{i=1}^{N}\delta_i.
\end{equation}
A value near zero says that positive and negative changes balance on average. It doesn't say that items were stable. If one item improves by .20 and another deteriorates by .20, \(L=0\) while neither item held still. To separate average stability from item movement and concentration, use the item instability (\(\instab\)) and worst-tail degradation (\(\wtd\)) summaries introduced by \textcite{zhang2026illusionRobustness}. Item instability averages the size of each change and ignores its direction:
\begin{equation}\label{eq:instability}
\instab=\frac{1}{N}\sum_{i=1}^{N}|\delta_i|.
\end{equation}
The same pair has \(\instab=.20\). Instability splits into a mean improvement \(F^+\) and a mean deterioration \(F^-\), each averaged over all \(N\) items rather than over the ones that moved. Their difference is the net change, \(L=F^+-F^-\), and their sum is instability, \(\instab=F^++F^-\). Reporting \(L\) and \(\instab\) together makes cancellation observable rather than hidden.

A mean can also hide concentration. For a prespecified tail fraction \(q\), order the paired changes from smallest to largest, writing \(\delta_{(j)}\) for the \(j\)th smallest, and average the worst \(q\) of them; at \(q=.1\) that's the worst-changing decile. Let \(m=\lceil qN\rceil\) be the number of items in the tail, rounding up. The sign is reversed so that a positive value means deterioration:
\begin{equation}\label{eq:wtd_main}
\wtd_q=-\frac{1}{m}\sum_{j=1}^{m}\delta_{(j)}.
\end{equation}
Worst-tail degradation still measures \mention{change}. A serious error repeated in both conditions has \(\delta_i=0\) and disappears from \(L\), \(\instab\), and \(\wtd_q\).

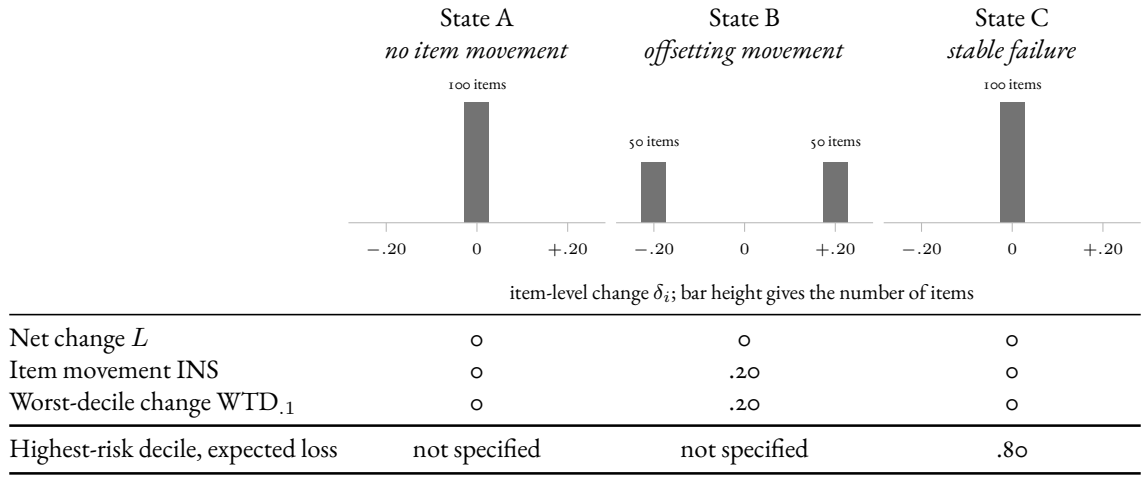
\begin{figure}[H]
\centering
\footnotesize
\setlength{\tabcolsep}{2pt}
\begin{tabular}{@{}l*{3}{>{\centering\arraybackslash}p{34mm}}@{}}
& \textbf{State A} & \textbf{State B} & \textbf{State C} \\
& \textit{no item movement} & \textit{offsetting movement} & \textit{stable failure} \\[3pt]
& \begin{tikzpicture}[x=1mm,y=1mm,font=\tiny]
\useasboundingbox (-17,-5.2) rectangle (17,19.5);
\fill[black!55] (-1.6,0) rectangle (1.6,16);
\node[anchor=south] at (0,16.5) {100 items};
\draw[gray!55] (-17,0) -- (17,0);
\draw[gray!55] (-12,0) -- (-12,-1.2);
\draw[gray!55] (0,0) -- (0,-1.2);
\draw[gray!55] (12,0) -- (12,-1.2);
\node[anchor=north] at (-12,-1.6) {$-.20$};
\node[anchor=north] at (0,-1.6) {$0$};
\node[anchor=north] at (12,-1.6) {$+.20$};
\end{tikzpicture} & \begin{tikzpicture}[x=1mm,y=1mm,font=\tiny]
\useasboundingbox (-17,-5.2) rectangle (17,19.5);
\fill[black!55] (-13.6,0) rectangle (-10.4,8);
\fill[black!55] (10.4,0) rectangle (13.6,8);
\node[anchor=south] at (-12,8.5) {50 items};
\node[anchor=south] at (12,8.5) {50 items};
\draw[gray!55] (-17,0) -- (17,0);
\draw[gray!55] (-12,0) -- (-12,-1.2);
\draw[gray!55] (0,0) -- (0,-1.2);
\draw[gray!55] (12,0) -- (12,-1.2);
\node[anchor=north] at (-12,-1.6) {$-.20$};
\node[anchor=north] at (0,-1.6) {$0$};
\node[anchor=north] at (12,-1.6) {$+.20$};
\end{tikzpicture} & \begin{tikzpicture}[x=1mm,y=1mm,font=\tiny]
\useasboundingbox (-17,-5.2) rectangle (17,19.5);
\fill[black!55] (-1.6,0) rectangle (1.6,16);
\node[anchor=south] at (0,16.5) {100 items};
\draw[gray!55] (-17,0) -- (17,0);
\draw[gray!55] (-12,0) -- (-12,-1.2);
\draw[gray!55] (0,0) -- (0,-1.2);
\draw[gray!55] (12,0) -- (12,-1.2);
\node[anchor=north] at (-12,-1.6) {$-.20$};
\node[anchor=north] at (0,-1.6) {$0$};
\node[anchor=north] at (12,-1.6) {$+.20$};
\end{tikzpicture} \\[2pt]
& \multicolumn{3}{c}{\scriptsize item-level change \(\delta_i\); bar height gives the number of items} \\
\midrule
Net change \(L\) & 0 & 0 & 0 \\
Item movement \(\instab\) & 0 & .20 & 0 \\
Worst-decile change \(\wtd_{.1}\) & 0 & .20 & 0 \\
\midrule[\heavyrulewidth]
Highest-risk decile, expected loss & not specified & not specified & .80 \\
\bottomrule
\end{tabular}
\caption{Three item-level states with the same net change, constructed for illustration rather than measured. Horizontal position gives each item's change \(\delta_i\) and bar height the number of items. The change statistics separate state B from the other two and don't separate A from C, whose ten highest-risk items carry an expected loss of .80 in both conditions. A source report that publishes only \(L\) leaves a downstream claim about final defects unanswerable}\label{fig:states}
\end{figure}

The distinction is visible without simulation. Fig.~\ref{fig:states} constructs three states of 100 items, all with \(L=0\). State A has no item movement. State B combines 50 improvements of .20 with 50 deteriorations of .20. State C has no movement either, but ten of its items carry an expected loss of .80 on the target outcome in both conditions, so it matches A on every change statistic and differs on absolute risk.

The states license different claims. State A supports stability on the observed items. State B defeats item stability despite the stable mean. State C shows why stability itself isn't safety: an unchanged serious failure remains serious. No estimator can recover which state obtained from \(L\) alone. The target claim determines which distinctions are required. A claim about answer-preserving context variation needs paired changes and their concentration; a deployment claim about final defects needs an absolute loss after the relevant review procedure.

State B isn't only a possibility. In one released design that adds task-irrelevant context to benchmark items \citep{zhang2026illusionRobustness}, the median absolute signed change across 32 model--benchmark comparisons was .0068, while the median item instability, the mean absolute item-level change, was .0678. That's under one percentage point of net movement alongside nearly seven percentage points of mean absolute item-level movement. Those comparisons share items within benchmark and lineage across models, so they're one crossed dataset rather than 32 independent replications, and what they show is how that dataset behaves rather than how evaluations behave in general.

An absolute target loss needs its own tail, and there are two of them. Each case carries a loss on the target outcome that varies with residual response, scoring, and outcome randomness, and averaging over that randomness gives the case's expected loss. Ordering cases by expected loss gives a case-risk tail, which picks out the cases expected to be risky. Ordering single realized draws gives a different tail, which also reflects residual bad luck in otherwise moderate-risk cases. A projectibility declaration, set out in Section~\ref{ssec:declaration}, has to say which object the decision concerns. Neither tail should be smuggled into a behavioural-change statistic.

\subsection{A known-truth demonstration}\label{ssec:released}

The three states above are logical possibilities. Whether an estimator can tell them apart is a separate question, and it can be settled under a known truth rather than argued. The empirical companion to this paper builds the scenarios below, pins its source version and data hashes, and provides the estimator code \ifanon in a companion repository whose URL is withheld for blind review\else at \url{https://github.com/BrettRey/benchmark-inference-composition}\fi.

In the stable-poor scenario, the same low-performing items have unchanged response probabilities under baseline and altered conditions. Latent worst-tail change is zero, while their worst-decile conditional expected loss is .80. The response-half estimate of the change tail is .0013; the cross-fitted case-risk tail, computed by reversing the halves and combining, is .7911. The disagreement persists without any ambiguity about the truth, because the scenario was built to contain it. That's what makes it a check on the estimators rather than a finding about the world: it establishes that a change tail and a case-risk tail can lie far apart while each estimator recovers its own target, not that such cases are common.

Selection is the second thing a known truth exposes. Choosing the worst observed tail and estimating it on the same finite trials inflates its apparent magnitude, a Type M error in the sense of \textcite{gelman2014types}. Splitting the trials, so that one half selects the items and the other estimates their changes, removes that particular inflation while leaving a different estimand: the latent effect of a noisily selected set rather than the tail selected from unobserved latent effects. A report should name which one it gives.

This section makes no claim about how often the states occur in released evaluations, which would need a different design and evidence this paper doesn't present. Its point is architectural, and it holds whatever the field's aggregates turn out to look like: an evaluator who publishes only a total leaves a downstream evaluator unable to inspect whether the source failures line up with the features that define its target.

\subsection{Evidence preservation at the interface}\label{ssec:preservation}

A projectibility audit imposes an information requirement on source reports. The developer needn't publish every conceivable statistic, but it should retain the resolution at which target-relevant distinctions can later be checked: item identifiers or auditable equivalents, item-level outputs and scores, template or source clusters, prompts and tool conditions, repeated-response structure, scorer decisions, model build and lineage, and registered alterations. A domain profile is better than a total when domains matter, but it can still hide item-level reversals and stable failures within each domain. The same requirement arises wherever a report compresses a judgment: in safety evaluation, pass/fail labels obscure whether a failure came from a capability limit, an ambiguous policy, an instruction conflict, or an unstable evaluator judgment, and separating those is what lets a later claim rest on the right one \ifanon\citep{anonSelfAdversarialPragmatics}\else\citep{reynolds2026adversarialpragmatics}\fi.

The requirement isn't maximal disaggregation for its own sake. Raw data can be sensitive, proprietary, or too large to release. The relevant standard is whether a declared downstream claim can be audited. Access controls, secure evaluation environments, independent auditors, or sufficiently detailed derived data may answer the need. What doesn't answer it is an aggregate from which several target-relevant states are observationally indistinguishable.

The legal example illustrates the interface. A high legal-domain mean doesn't reveal whether errors concentrate on questions requiring current authority or on outputs with plausible but unsupported citations. A local evaluator needs those distinctions to decide what to sample and score. If the developer can report item changes under retrieval-like distractors, citation-related errors, and build provenance, that evidence may inform the firm's design. It still doesn't replace the firm's observations of its own requests and review procedure.

\section{What the Non-Composition Principle Changes for AGI Evaluation}\label{sec:agi}

Multidomain AGI evaluations make the composition problem unusually acute. The label \mention{general} invites a conclusion that reaches beyond every fixed set of tasks, systems, and conditions used to construct the score. Adding domains increases coverage, but coverage and inferential reach aren't the same. Before a total or profile is read as evidence of general capability, three questions have to be separated: whether its components share a scale, whether the permitted tradeoffs are acceptable, and whether the resulting description projects to the target claim.

\subsection{Commensurability, compensability, and projectibility}\label{ssec:three_questions}

\term{Commensurability} asks whether differences on component scores have a common interpretation that makes addition meaningful. Putting a legal score and a quantitative score on a common scale from zero to one doesn't show that a .05 increase represents the same amount of improvement in both. Equal weights don't solve the problem; components with different variances and covariances can make unequal statistical contributions even under equal nominal weights \citep[305--308]{brennan2001generalizability}.

\term{Compensability} asks whether gains on one component may offset failures on another for the interpretation or use at issue. A weighted sum permits such substitutions over the reported ranges. That may be acceptable for a descriptive index. It's unacceptable when the target imposes a bottleneck or red line. In the legal case, a high quantitative score doesn't compensate for an unresolvable citation, and success on routine termination-clause research doesn't compensate for failure on reasonable-notice requests if both uses are to be authorized. Value models can represent accepted tradeoffs, but the weights are range-dependent preferences rather than context-free measures of domain importance \citep{KeeneyRaiffa1993}.

\term{Projectibility} asks whether the aggregate supports a specified claim beyond the cases and conditions used to construct it. Even a commensurable, defensibly weighted index may provide little warrant for a projection to a new task, system, or period. Conversely, a score can predict a bounded criterion without being an interpretable measure of a single latent capability.

\subsection{From taxonomy to decision: four inferential roles}\label{ssec:chc_roles}

The Cattell--Horn--Carroll (CHC) framework used in some AGI proposals can enter an evaluation at four points \citep{carroll1993human,mcgrew2009chc,schneider2018chc,hendrycks2025agi}. It can organize benchmark content, so that a battery doesn't consist entirely of one familiar task format, without establishing that artificial systems reproduce the human covariance structure that motivated the taxonomy. Covariation among task scores in a declared population of systems can be summarized by a factor model. Those scores can be interpreted as capacities with a process or nomological meaning; a network of expected task, process, and external relations can make that interpretation more substantive, but the relations still need testing in the artificial-system population \citep{cronbach1955construct,freiesleben2026establishing}. The score or profile can then predict an external criterion or guide a decision. These aren't four kinds of evidence for one conclusion. They're successive roles joined by further inferences, and evidence for one doesn't warrant the next.

Each role carries its own requirement. As a \term{description}, a score reports a declared operation on the construction sample, and accuracy is largely calculation, sampling, and scoring. As a \term{measure}, it represents a capability and needs content, structural, process, and external evidence appropriate to that interpretation \citep{messick1995Validity,borsboom2004conceptValidity}. Psychometric theory already separates evidence about the processes items elicit from evidence about relations among scores and external criteria, and strong support for one needn't supply the other \citep[180]{embretson1983construct}. As a \term{predictor}, the score needs target-matched out-of-sample performance at the independent unit the claim names. As a \term{decision input}, it also needs a defensible relation to consequences, feasible policies, and the people who bear their effects.

Across 591 nominally distinct models and 12 tests, \textcite[4--5]{ilicGignac2024cognitiveCapabilities} report a positive manifold and a strong general factor, though the four-factor hierarchy beneath it wasn't interpretable. That supports the structural description for one sampled model--task matrix. It doesn't establish the CHC decomposition, its construct interpretation, or its transport to later systems. \Textcite{krakauer2026riseFall} shows the last isn't hypothetical. On a four-benchmark battery the first principal component holds 92\% of variance in 2023--2024 and 64\% once inference-time reasoning models arrive, with the ratio of the first eigenvalue to the second falling from 15:1 to 1.8:1. He reports the analysis as preliminary and notes that four benchmarks can't statistically confirm a second factor, but a structure that moves this much within the sampled period isn't a fixed property of the systems.

Comparability across model families needs separate evidence that items and score relations keep their meaning \citep{meredith1993measurementInvariance}. \Textcite{jung2026psychometric} give a direct warning about ordering systems. Ranking language models by their scores on human psychometric instruments didn't recover the ranking by behaviour on matched downstream tasks, and for sexism and racism the two ran opposite. Instruments that were moderately reliable under item and prompt variation still ordered the systems wrongly.

A common error is to validate one role and write as though the others follow. High test--retest reliability supports score stability under the repeated protocol, not a capability ontology, and a factor loading records association in a fitted model, not an internal mechanism. Theorized benchmark construction can strengthen the content and interpretation steps \citep{barmanEtAl2024understanding} and still leaves the downstream projections to be shown. The non-composition principle isn't an objection to theorized benchmarks. It specifies the additional work their results require downstream, and projectibility is then assessed for each move rather than attributed to the total as a prestige property.

\subsection{Why accumulated coverage doesn't entail unrestricted generality}\label{ssec:coverage}

Adding domains broadens the benchmark's source description, but it doesn't by itself show how the result extends beyond those domains. A finite battery can define an index over its construction sample and support bounded generalizations when its item-generating process, sampling assumptions, and uncertainty are defensible. The stronger claim begins when that index is treated as evidence of capability across tasks and conditions the process didn't represent. At that point the evaluator needs evidence for the rule of extension, not a larger domain count.

This is the Goodman point in practical form. Passing tasks in law, mathematics, coding, and spatial reasoning fits a general-capability hypothesis. It can also fit a collection of narrower hypotheses under which performance depends on familiar interfaces, training overlap, static inputs, short horizons, or the absence of tool and user interactions. The hypotheses agree on the battery and diverge elsewhere. Domain count alone doesn't select among them. Evidence has to vary the suspected conditions or sample the target cases.

Cross-domain covariation is stronger evidence than a domain count, and the positive manifold reported above supplies it for that sampled matrix without establishing unrestricted reach. Separate studies might support transfer from benchmark mathematics to unseen textbook problems, from coding exercises to repository-level bug fixes, and from supplied-text law questions to bounded research requests. Their union says more than any one, but without evidence relating those targets it doesn't establish arbitrary tasks, domain interactions, later systems, or novel tool environments.

Bounded evidence can accumulate legitimately. If the cases reached by one projection are sampled by the next, interface assumptions are tested, and dependence and uncertainty are preserved, the chain can extend. A system's benchmark profile might predict target-task outcomes across independent builds; target-task outcomes might predict final workflow results across sites; those results might remain stable in chronological holdouts. The conclusion is as broad as the composed path and no broader. The non-composition principle blocks an automatic inference, not empirical progress.

\section{Procedure and Evidential Responsibility}\label{sec:procedure}

A projectibility audit is useful only if it changes what evaluators record and test. Its procedure begins before calculation, with a sentence that can be contradicted by observations and a typed description of the source and target. It ends neither with a benchmark report nor with authorization. Later outcomes and material system changes reopen the links they bear on.

\subsection{Who can supply the evidence}\label{ssec:responsibility}

The evidential burden is distributed because no single actor observes the whole chain. A developer can describe the source evaluation in detail. It can identify the exact model build and declared lineage; publish or provide controlled access to items, prompts, tool settings, scoring rules, item-level outputs, and repeated-response structure; document which templates or sources were held out; report registered alterations and known failures; and state which product changes invalidate the result.

Psychometric analyses can improve the quality and interpretation of the instrument itself, as work on item-response methods for NLP benchmarks illustrates \citep{bachmannEtAl2024psychometrics}. Those analyses remain conditional on the systems, items, and score interpretations studied.

A deployer observes different facts. Only the law firm in the running example can define which requests it receives, sample them from its logs, prepare local authority lists, observe its lawyers' review, compare the policies it can adopt, and record whether a defect remains internal or reaches a client or court. A hospital, school, or public agency would have corresponding task, workflow, exposure, and outcome records. The inability of a developer to foresee those details narrows the developer's warranted claim. It doesn't turn a generic benchmark into evidence about every use.

Neither developers nor deployers can validate the whole chain alone. The deployer needs notice of model, safety-layer, retrieval, or API changes; enough configuration information to freeze the tested object; and access to outputs at the resolution required for local scoring. The developer needs feedback about failures that source evaluation didn't represent. Independent evaluators may test both sides, but they can't infer proprietary lineage or local workflow conditions from a public product label. Where information remains unavailable, the corresponding projection remains conditional or unsupported rather than being filled by an assumption of continuity.

Responsibility is divided but not dissolved. Developers are responsible for the interpretations and uses they advertise and for foreseeable interface failures; deployers remain responsible for the decisions they make and the evidence available only in context. A claim that no party can test may still be rhetorically attractive, but the absence of an evidential owner is itself a defect in the argument.

\subsection{The projectibility declaration}\label{ssec:declaration}

For confirmatory use, the evaluator should complete a \term{projectibility declaration} before opening the final holdout. Exploratory work can use the same fields, but it should reserve new evidence for confirmation. The declaration has thirteen parts.

\begin{enumerate}
\item \textbf{Write one source--target claim.} State what was observed and the new cases, interpretation, or outcome claimed. Replace \mention{legal capability transfers to practice} with a statement such as: final memos produced by the frozen application and registered review procedure will have a post-review substantive-defect probability below 3\% on Ontario termination-clause requests admitted by the written rule.

\item \textbf{Type the source and target.} Record the object, population, conditions, outcome, and period at each endpoint. If the claim is a capability attribution, it isn't an empirical endpoint; record instead the observable relation the downstream link needs and the evidence that would test it. A shared label doesn't count as endpoint identity.

\item \textbf{Name the extension.} Specify whether the claim generalizes to further items, interprets a score as a construct, extrapolates to another task, transports across systems or sites, predicts a later period, explains an outcome, or links evidence to a policy. Composite claims should be split until each boundary is visible.

\item \textbf{List plausible defeaters.} Name actual differences that could change the result: live rather than supplied retrieval, added irrelevant cases, another database index, a different reviewer instruction, a new model lineage, conflicting authority, or a later work period. The list is justified by prior failures, domain knowledge, process evidence, and stakeholder concerns, not by a generic field headed \mention{context}.

\item \textbf{Assign evidence and ownership.} For each assumption or defeater, state the observation that bears on it and who can produce that observation. If no actor can observe whether defects reach clients, a claim about client consequences isn't ready for confirmation.

\item \textbf{Match the sampling and holdout unit to the claim.} Keep complete templates out for a template claim, client-file clusters out for a request claim, alteration types out for a context claim, lineages out for a system claim, offices out for a site claim, and later periods out for a temporal claim. Repeated outputs estimate variation within a request--condition pairing; they don't create new requests, systems, or periods.

\item \textbf{Define the observation and loss.} State exactly what one row represents and what is scored. Distinguish a request, a generated draft, a lawyer review, a final memo, a system build, and a work period. Keep ordinal hazard classes separate unless a defensible cardinal scale exists. Identify red-line outcomes that may not be averaged away.

\item \textbf{Register support, defeat, and unresolved conditions.} Give the estimate, uncertainty requirement, minimum sample or tail count, multiplicity treatment, and any fallback claim before examining confirmatory results. An inconclusive result remains inconclusive; it isn't evidence of equivalence or robustness.

\item \textbf{Audit the interfaces.} For every adjacent link, ask whether the upstream target is the downstream source in object, population, conditions, outcome, and period; whether assumptions are compatible; and how dependence and uncertainty are propagated. Classify the relation as composition, convergence, replacement, or no evidential connection.

\item \textbf{Separate the empirical claim from the decision.} List feasible policies, affected parties, observed and prospective consequences, costs, hard constraints, and update triggers. State which empirical results enter the choice and which values determine the boundary. A validity judgment doesn't select a policy by itself.
\end{enumerate}

Table~\ref{tab:declaration_fields} gives a compact record. It carries more than Equation~\ref{eq:node} because a declaration covers a projection rather than a node: the five node fields recur, but the record also has to state which endpoints are being joined, what the edge assumes and what evidence bears on it, and what the decision then requires. The final column prevents the declaration from becoming a set of decorative headings: every entry resolves into a value, sampling rule, or named document.

\begin{table}[tbp]
\centering\small
\caption{Minimum projectibility-declaration record}\label{tab:declaration_fields}
\begin{tabular}{@{}>{\raggedright\arraybackslash}p{0.22\linewidth}
                    >{\raggedright\arraybackslash}p{0.31\linewidth}
                    >{\raggedright\arraybackslash}p{0.38\linewidth}@{}}
\toprule
Field & Required statement & Inspectable form in the legal example \\
\midrule
Source observations & What was directly recorded & Item responses under the benchmark protocol; or draft and final defect scores on named request identifiers \\
Target & Cases or interpretation reached & Requests satisfying the Ontario termination-clause inclusion rule during the declared period \\
Object & The tested system or workflow, identified & Model build, retrieval index, prompt stack, drafting template, citation validator, and review procedure \\
Unit and population & What one observation is and how cases enter & One logged research request; one request selected per client file; 200 later files per type \\
Conditions & Settings under which the object was operated & Decoding settings, retrieval parameters, the firm's licensed database, the scorer, and the written review instruction \\
Outcome and loss & What counts as failure or cost & Unresolvable citation, unsupported proposition, omitted controlling authority, final defect indicator, and lawyer minutes \\
Period & Version and span each endpoint covers & Benchmark evaluation date at the source; the declared six-month work period at the target \\
Defeaters & Differences expected to matter & Irrelevant retrieved cases, request type, index update, reviewer change, and new build \\
Evidence design & Holdout or intervention matched to each difference & Later client-file clusters; answer-preserving retrieval alteration; separate request-type estimates \\
Decision rule & Support, defeat, and inconclusive conditions & Registered exact bound, 3\% tolerance, red-line rule, and prespecified narrower policy \\
Interface & Relation to each adjacent claim & Local study replaces rather than confirms the benchmark-to-use projection \\
Transmission & Modifiers, dependence, and uncertainty carried across any join & Request type retained as a stratifier rather than pooled; client-file clustering preserved in the uncertainty analysis \\
Update trigger & Event that reopens the claim & Model, index, prompt, request type, office, or review-procedure change; adverse incident \\
\bottomrule
\end{tabular}
\end{table}

\subsection{Designs matched to common projections}\label{ssec:matched_designs}

The declaration changes the statistical design because the independent unit follows the claim. To generalize from observed to unobserved responses under the same item rules, hold out item or template clusters rather than random response rows. To extrapolate from benchmark questions to professional tasks, sample external tasks by a written rule and score them against independently prepared criteria. To claim robustness to a new alteration family, keep one whole alteration type out of tuning.

To claim transfer to another system, evaluate a genuinely different build or lineage; repeated drafts from one build only make the estimate for that build more precise. To claim transfer to another office, run the complete task and review procedure there. To predict later performance, use a chronological holdout at the stated horizon.

Construct claims need another design. A taxonomy needs content argument. A factor interpretation needs a declared task and system population, adequate variation, and checks that the fitted relations aren't artefacts of a narrow item set. A nomological claim needs predicted convergent, discriminant, process, and criterion relations. A causal explanation needs interventions or other identification conditions that discriminate plausible rivals. More benchmark items can improve precision within the tested operation while leaving each of those obligations untouched.

Predictive claims make an especially common unit error visible. Suppose the question is whether benchmark profiles predict legal-research defects across systems. The independent units are systems or lineages, each with a benchmark profile and a defect estimate on the same held-out requests. With one system, its profile doesn't vary across requests and can't predict which request fails. A request-level model instead needs request-level predictors. Fitting the system profile to hundreds of request rows manufactures a sample size by repeating a constant.

The same discipline applies to tails and groups. A worst-decile estimate selected after examining noisy item effects is subject to selection inflation; response splitting, cross-fitting, or a hierarchical model can address that problem for a stated estimand. None addresses transport to a new task type or system. Sparse group samples can leave a group-specific claim unresolved; a pooled estimate isn't automatically a licensed substitute. If the decision constrains the maximum defect tail across two request types, uncertainty has to account for selecting that maximum.

\subsection{Reporting and updating}\label{ssec:reporting}

A projectibility report should preserve continuous estimates and uncertainty even when a policy uses a boundary. The supported, defeated, or unresolved label summarizes the registered relation; it doesn't replace the estimate. Reports should also distinguish raw descriptive quantities from estimates that are null-referenced (subtracting the value the statistic takes under a simulated no-difference condition), split-sample, cross-fitted, or model-based, and state what each targets. A visually similar number can answer a different question.

After authorization, the evidential record continues. A shadow phase can run the application beside ordinary work without allowing it to determine the memo sent onward. The firm records draft defects, corrections, review time, final defects, and any incident. Continued use adds exposure and outcome records. A model update, retrieval-index change, revised prompt, new request type, changed review procedure, or shift in observed failures reopens the relevant links.

Frequent model releases don't require every part of an evaluation to be repeated from scratch. An update reopens the links whose object, conditions, or assumptions it changes. Task definitions, inclusion rules, reference standards, scoring procedures, and evidence about unchanged parts of the workflow can often be reused; claims about model outputs require new evidence when the tested build changes materially.

Version pinning, change logs, regression tests, rotated holdouts, and shadow evaluation can reduce the cost, while the extent of retesting should reflect the stakes, the opacity of the change, and the scope of the authorized claim. If a provider doesn't permit a build to be frozen or supply enough information to determine what changed, the result is a narrower or conditional authorization. Sometimes rapid model turnover makes the proposed use impractical; it doesn't make evidence about an earlier model applicable by default.

Monitoring isn't a substitute for predeployment evidence. Allowing a high-stakes system to produce consequences in order to learn whether it's safe simply relocates the cost of uncertainty. Nor is predeployment testing a substitute for monitoring, since later users, tasks, and systems can differ. The two answer temporal links on opposite sides of authorization.

The audit can end with a narrow claim. Sparse or mismatched evidence doesn't imply that the system is generally unreliable; it implies that a broader claim lacks support. Conversely, a successful local study doesn't prove general capability. Precision about scope is an empirical result, not a rhetorical concession.

\subsection{Proportional implementation}\label{ssec:proportional}

The framework scales down; the worked design isn't an entry price. A firm unable to run the full confirmatory design can use developer and independent evaluations for source-side characterization while conducting a lighter local study: a modest audit of closed matters against the best available file record, a shadow phase, and incident logging. Such evidence supports only a conditional, narrower authorization, with omitted-authority risk, rare failures, and unsampled conditions recorded as unresolved. The scope can widen as relevant evidence accumulates. The cost of a broad authorization is high because a broad authorization asserts a great deal, which is the argument rather than an obstacle to it.

Populating endpoint records and checking their fields for correspondence are largely clerical tasks, and are therefore amenable to machine assistance. Identifying plausible defeaters and judging the adequacy of a reference standard aren't clerical; those are where substantive expertise and much of the cost remain.

\section{Limitations}\label{sec:limitations}

Projectibility isn't a complete theory of induction. The framework requires evaluators to identify plausible defeaters, but observations never determine that list by themselves. Domain theory, prior failures, causal knowledge, institutional experience, and affected stakeholders all shape which differences are investigated. Those sources can be incomplete or contested. A projectibility audit makes the commitments visible; it doesn't guarantee that every important difference has been anticipated.

The same underlying failure can receive either diagnosis, depending on how finely the endpoints are described. Under a coarse schema it may appear as a failure of warrant transmission; adding a field for the relevant distinction can turn it into an endpoint mismatch. The classification is therefore representation-relative, though not arbitrary. The two diagnoses are detected and repaired differently, and the audit records which distinctions its declared schema makes visible rather than claiming to recover a description-independent division in the world.

Because the statuses are claim-relative, an evaluator can rescue a failing broad claim by narrowing it until it passes. Prospective declaration and independent scrutiny limit such rescue without eliminating it, and the practical question is whether the narrow claim remains useful enough to justify the proposed policy. Completing the declaration doesn't confer warrant either. Every field can be populated over a poor reference standard, implausible invariance assumptions, or evidence too weak for the declared use.

The interface requirements aren't a replacement for specialized methods. Causal projections require causal identification; psychometric interpretations require appropriate measurement models and construct evidence; predictive claims require target-matched validation; and decisions require normative and institutional justification. The framework's contribution is to keep those analyses from being joined merely because their outputs share a label. Nor is the formalism a model of feedback dynamics. It represents paths through a chain and can reopen links as systems, users, and workflows change, but it doesn't explain those adaptations.

The legal results are constructed. They demonstrate how a complete audit assigns different statuses, not how any actual assistant performs. The simulations establish that the estimands come apart under a declared generating process. They don't establish how often that happens in released evaluations, don't validate the projectibility framework against deployment outcomes, and don't show that any benchmark quantity predicts another target. The framework itself should be studied prospectively: evaluators could compare decisions, failure detection, and claim revision with and without an explicit interface audit.

Some evidence will remain inaccessible. Proprietary training overlap can make model lineages uncertain; privacy and privilege can limit release of local cases; rare harms can make direct estimation impractical; and important future conditions may be unforeseeable. The response isn't to infer continuity by default. Evaluators can use secure audits, partial identification, stress tests, sensitivity analysis, and narrower authorization, while recording which links remain conditional.

Finally, the procedure can be expensive. Sampling target tasks, preparing independent reference standards, testing complete workflows, and retaining outcome records require expertise and time. Those costs should be compared with the value and stakes of the proposed use. A low-consequence drafting aid may justify a lighter evidential standard than a system whose outputs enter court filings without independent verification. Claim-relative standards aren't evidential relativism; they connect the cost of being wrong to the strength and kind of support required.

\section{Conclusion}\label{sec:conclusion}

Validity-centred AI evaluation has made an essential correction: validity belongs to a proposed interpretation and use, not to a benchmark in isolation. This paper takes up what that correction leaves to the analyst. Its object is the joins between separately produced claims, and it gives them a representation and an audit. A benchmark result can warrant several bounded inferences without warranting the chain assembled from them. A capability attribution may support one link, but it doesn't itself specify the downstream cases, criterion, or conditions the next link needs.

Goodman shows why source fit alone can't choose among rival extensions, and argument-based validity locates each extension among explicit claims and assumptions. Projectibility concerns whether one such bounded extension is warranted, and the audit asks two questions in order. Do links presented as adjacent meet in object, population, conditions, outcome, and period, or does the join need a separately warranted bridge? If they meet, do their assumptions and effect modifiers stay compatible, and are dependence and uncertainty carried across? Failure at the first leaves the composition ill-typed as stated. Failure at the second leaves a well-typed chain unwarranted.

The legal example shows the practical consequence. A developer's score on supplied-text questions and a firm's local study of reviewed memos can each be sound while remaining parallel, and direct local evidence may replace the benchmark-to-use projection rather than confirm it. The local study can support mandatory-review use on one request type, defeat another, and leave a changed retrieval condition unresolved, without settling what the base model can do. A later build, another office, or another request type reopens a different link.

Aggregation raises the same problem inside a single report, because joins require information at the resolution of the downstream claim. One stable mean can conceal balanced changes, concentrated deterioration, or stable serious failures. For AGI evaluation, broader domain coverage can likewise improve the source description without establishing commensurability, acceptable compensation, a common capability, or unrestricted reach.

Developers should report what their evaluations observed and preserve the resolution a later audit needs; deployers should sample the tasks, workflows, and consequences available only in context. Bounded projections accumulate when endpoints align and warrant transmits across each join. The unit of warrant is then not the benchmark score but the declared, revisable path from observation to use, and its conclusion is as broad as that path and no broader.

\appendix

\section{Interface Audit Questions}\label{app:interface_questions}

For each pair of adjacent links, the evaluator can use the following questions as a compact composition audit. The first eight run the requirements of Section~\ref{ssec:interfaces}, with dependence and uncertainty taken separately; the last three concern a proposed composition as a whole and have no counterpart there.

\begin{enumerate}
\item \textbf{Do the endpoints denote the same object?} If the upstream target is a base model and the downstream source is a retrieval application or reviewed workflow, what evidence links them? If either endpoint is named as a capability, which tested score--criterion relation stands in for it?
\item \textbf{Are the cases aligned?} Does the downstream sampling rule admit the cases reached upstream, with the same exclusions, grouping, and target weights?
\item \textbf{Is the outcome continuous?} Does the first link deliver the variable the second consumes, or has accuracy become a factor score, defect probability, reviewed outcome, or consequence without a tested mapping?
\item \textbf{Are operating conditions compatible?} Which prompts, tools, databases, reviewers, and sites differ, and which observations test those differences?

\item \textbf{Do the periods align?} Which evaluation dates and claimed work spans does each endpoint cover, and is the difference tested by a chronological holdout rather than assumed continuous? Build and index identity belong to the object question; a retest trigger conditions future authorization and doesn't answer a present difference.
\item \textbf{Are assumptions inherited consistently?} Does the second link rely on an invariance, independence, or process claim contradicted or left open by the first?
\item \textbf{Is the evidence as independent as claimed?} Which items, templates, scorers, training runs, model lineages, or development choices are shared?
\item \textbf{Has uncertainty been carried through?} Are estimates, selection procedures, and unresolved qualitative assumptions propagated rather than replaced by point conclusions?
\item \textbf{Is this actually composition?} Would the downstream claim be supported if the upstream study didn't exist? If so, the relation may be replacement or convergence rather than composition.
\item \textbf{What observation would defeat the composed claim?} A chain with no stated defeater hasn't yet exposed its empirical content.
\item \textbf{Does the conclusion stop at the reached target?} Successful links warrant the declared path, not unspecified tasks, systems, or times beyond it.
\end{enumerate}

A completed audit needn't show that every interface is secure. Its first function is diagnostic. A missing link can lead to a new study, a narrower claim, a conditional conclusion, or rejection of the proposed use. What it can't legitimately produce is scope by accumulation: several passed evaluations don't add up to a claim whose interfaces were never tested.

\clearpage
\printbibliography

\end{document}